\PassOptionsToPackage{capitalize,noabbrev,nameinlink}{cleveref}                                                                                                                                                 
\PassOptionsToPackage{usenames,dvipsnames}{color}

\documentclass[nonacm,10pt]{acmart}
\usepackage{times,amsmath,epsfig}
\usepackage{enumitem}

\newcommand{\paperTitle}{Question Answering via Web Extracted Tables and Pipelined Models}

\usepackage{amsmath}
\usepackage{amssymb}
\usepackage{boxedminipage}
\usepackage{xspace}
\usepackage{tabularx}
\usepackage{balance}  
\usepackage[hyphenbreaks]{breakurl}
\usepackage[font={small}]{caption}
\usepackage{graphicx}
\usepackage{subfig}

\usepackage{graphicx}
\usepackage{subfig}
\usepackage{soul}
\usepackage{pifont}

\usepackage{booktabs}
\usepackage[end]{algpseudocode}
\usepackage{algorithm}
\usepackage{hyperref}

\newcommand{\mlbegin}{\shortstack\bgroup}
\newcommand{\mlend}{\egroup}

\newcommand{\squishitemize}{
 \begin{list}{$\bullet$}
  { \setlength{\itemsep}{0pt}
     \setlength{\parsep}{3pt}
     \setlength{\topsep}{3pt}
     \setlength{\partopsep}{0pt}
     \setlength{\leftmargin}{1.95em}
     \setlength{\labelwidth}{1.5em}
     \setlength{\labelsep}{0.5em} } }

\newcounter{Lcount}
\newcommand{\squishlist}{
    \begin{list}{\arabic{Lcount}. }
   { \usecounter{Lcount}
        \setlength{\itemsep}{0pt}
        \setlength{\parsep}{3pt}
        \setlength{\topsep}{3pt}
        \setlength{\partopsep}{0pt}
        \setlength{\leftmargin}{2em}
        \setlength{\labelwidth}{1.5em}
        \setlength{\labelsep}{0.5em} } }

\newcommand{\squishend}{\end{list}}

\definecolor{todo-color}{rgb}{1,0,0}

\definecolor{comment-color}{rgb}{0.25,0.25,0.25}

\setcopyright{none}
\acmDOI{}

\acmISBN{}

\acmConference[]{}{}{}
\acmYear{}
\copyrightyear{}

\acmArticle{}
\acmPrice{}

\begin{document}
\title{\paperTitle}

\author{Bhavya Karki}
\author{Fan Hu}
\author{Nithin Haridas}
\author{Suhail Barot}
\author{Zihua Liu} 
\author{Lucile Callebert}
\author{Matthias Grabmair}
\author{Anthony Tomasic}
\affiliation{%
  \institution{Carnegie Mellon University}
  \city{Pittsburgh}
  \state{PA}
}

\renewcommand{\shortauthors}{B. Karki, et al.}

\begin{abstract}
In this paper, we describe a dataset and baseline result for a question answering that utilizes web tables. It contains commonly asked questions on the web and their corresponding answers found in tables on websites. Our dataset is novel in that every question is paired with a table of a different signature. In particular, the dataset contains two classes of tables: entity-instance tables and the key-value tables.
Each QA instance comprises a table of either kind, a natural language question, and a corresponding structured SQL query. We build our model by dividing question answering into several tasks, including table retrieval and question element classification, and conduct experiments to measure the performance of each task. We extract various features specific to each task and compose a full pipeline which constructs the SQL query from its parts. Our work provides qualitative results and error analysis for each task, and identifies in detail the reasoning required to generate SQL expressions from natural language questions. This analysis of reasoning informs future models based on neural machine learning.
\end{abstract}

\maketitle

\section{Introduction}
\label{sec:introduction}

State of the art question answering (QA) systems, such as Amazon Alexa or even Google's advanced search functionalities, produce answers by either retrieving text segments from a large repository of documents, or by retrieving entities from a large knowledge graph (e.g. Freebase) which need to be populated by hand or constructed from text using sophisticated NLP technology. At the same time, the the internet already contains much structured knowledge in the form of tables on websites. In this paper we work towards leveraging pre-existing tables on website by using them to produce answers to questions, which we translate into SQL queries that can be executed against the table. Current solutions to question answering of structured queries generally use neural models, which require very large datasets. We collected a dataset of questions and answers from tables. Since the collection is expensive it is not large enough to apply sequence-to-sequence (seq2seq)\cite{seq2seq} models that encode a natural language question and produce a structured query as the output. In this paper, we decompose the reasoning involved in constructing structured queries into a series of steps that translate a natural language questions the components of SQL queries, such as the SELECT clause, WHERE clause, and so on.

This approach offers a deeper understanding of the {\em semantic} and {\em syntactic} information required for generation of a structured query. This information is critical to guiding future
efforts in training set construction, merging with existing knowledge sources, and integration with structured sources. 
Tabular information includes regular tables, structured grids of information, lists of information, and property-value tables.
In summary, we make the our paper makes the following contributions:
\begin{itemize}
\item a decomposition of QA-from-tables into individual tasks and providing insights on each task, and
\item experimental results that deal with issues pertaining to limited data by extensive feature engineering and application of machine learning models in an intelligent manner.

\end{itemize}


\section{Related Work}

One of the earlier works in using structured data as knowledge sources~\cite{jauhar2016tables} focuses on exploring tables as semi-structured knowledge for multiple-choice question (MCQ) answering. The tables contain text in free form and are mostly general knowledge based facts. They contain answers to more than 9000 crowd-sourced questions in their base dataset AI2's Aristo Tablestore. The authors develop a feature-driven QA model that uses MCQs while performing fact checking and reasoning over tables. Similarly, our models heavily depend on features from text in tables to solve individual tasks.
Hayati et al.~\cite{hayati2018retrieval} use semantic and n-gram similarity for retrieval. They introduce ReCode, a sub-tree retrieval based framework for neural code generation using natural language. They use dynamic programming based methods to retrieve semantically similar sentences, then construct the syntax tree using n-gram of action sequences. The importance of semantic similarity features was noted and we experiment with incorporating semantic similarity features in our models. 
Cheng et al.~\cite{cheng2017learning} introduce a neural semantic parser that uses predicate-argument structures to convert natural language utterances to intermediate representations. These representations are then mapped to target domains. Similarly, we break down most of our questions into a projection column and conditional row to construct the structured query.
Waltz et al.~\cite{englishquery} builds a natural language QA system that produces answers a large relational database of aircraft flight and maintenance data using a system called PLANES. The system uses a number of augmented transition networks, each of which matches phrases with a specific meaning, along with context registers (history keepers) and concept case frames; these are used for judging meaningfulness of questions, generating dialogue for clarifying partially understood questions, and resolving ellipsis and pronoun reference problems. While we do not use exact 'phrase matching' to specific meanings, we employ machine learning models to infer similar meanings from the questions.
Chen et al. \cite{chenfair} proposes to tackle open domain QA using Wikipedia as the knowledge source. Their approach combines a search component based on bigram hashing and TF-IDF \cite{tfidf} matching with a multi-layer recurrent neural network model trained to detect answers in Wikipedia paragraphs.
We also employ used TF-IDF\cite{tfidf} for feature extraction used Wikipedia as a source for some of our data tables.
A study by Srihari et al.~\cite{iesq} shows that low-level information extraction tasks, like Named Entity tagging, are often a necessary
components in handling most types of questions. We also use Named Entity tagging as one of many information extraction features for classification at different steps of our pipeline.

WikiSQL and Seq2SQL (Zhong et al.)~\cite{zhong2017seq2sql} use a deep neural network for translating natural language questions to corresponding SQL queries. WikiSQL consists of 80654 hand-annotated examples of questions and SQL queries distributed
across 24241 tables from Wikipedia, which is an order of magnitude larger than other comparable datasets. The Seq2SQL model uses rewards from in-the-loop query execution over the database to learn a policy to generate the query. A pointer network limits the output space of the generated sequence to the union of the table schema, question utterance, and SQL keywords. Our dataset is similar WikiSQL but distinguishes between different types of tables. Because of our smaller size dataset we break down the approach into tasks that can be solved with per-task machine learning models. This approach essentially eliminates logical form errors that are prevalent in neural models.

\section{Data Collection}


\subsection{Collecting questions}

We created Amazon Mechanical Turk (AMT) tasks that asked workers about "5 questions you recently had where you searched for answers on the web"\footnote{\url{https://cmu-rerc-apt.github.io/QASdatacollection/examples.html}}. Workers were asked to generate questions from their past by going through their browser history, as opposed to other generation methods \cite{pasupat2015compositional, jauhar2016tables} where workers are primed to ask questions based on a given table. We did \textit{not} limit workers to questions that have an answer in tabular form to facilitate realistic questions.
Each worker was asked for 5 triples composed of (i) a natural language question, (ii) the URL of a page that contained the answer to the question, and (iii) the answer to the question. Workers were paid \$1.75 per question. In a pre-study we observed that workers generated a large amount of data about sports and weather, leading to bias in data collection. We therefore amended the task description to ban weather and sports questions.

\subsection{Collecting tables and SQL queries}

Collecting tables that contain the answer to our questions and corresponding SQL queries requires some expertise and skill (i.e. identifying tabular data, extracting a table and writing an SQL query). We hence recruited several students from a Computer Science Department. 
For each new question (previously collected through AMT), 1) search the web for a page with tabular data containing the answer. 2) Extract the relevant table from that page. Workers used two different tools to extract tables: import.io\footnote{Commercially available at: \url{https://www.import.io}} and SmartWrap \cite{gardiner2015smartwrap}). 3) Write an SQL query to extract the answer from the table. The costs associated with this elaborate, but necessary, process limited the size of our eventual dataset.
If any of the three steps failed for some reason, the question was discarded.

Upon examining the results, we noted that in many cases the SQL query substituted a word in the question with a related word in the WHERE clause of the query to insure that the LIKE operator would work. For example, "Who is the husband of Whoopi Goldberg?" would result in \verb|WHERE| \verb|value| \verb|LIKE| \verb|"spouse"| because a row in the table was identified using the word spouse.
This SQL query would require the system to understand some relationship between the concept of husband and the concept of spouse. 
We capture this issue by introducing a word embedding proximity operation \cite{bordawekar2016enabling} $\sim$. This operator measures the distance of every value in the column with the given value using the word embedding. Any value with a distance under a fixed threshold is considered a match.  Since husband and spouse are near each other in a typical word embedding\footnote{We used word vectors pre-trained on Google news available at \url{https://code.google.com/archive/p/word2vec}}, rewriting the clause to \verb|WHERE| \verb|value| $\sim$ \verb|husband| greatly simplifies the learning process, since this logical form only requires learning which token or tokens to copy from the natural language question to the clause. This form in a sense moves the concept mapping problem to the implementation of the $\sim$ operator.


\subsection{Dataset Analysis}

We collected 341 questions with associated table, SQL query, and expected answer. We divided the dataset into training set (238 examples), dev set (64 examples) and test set (39 examples). Each question is unique but some questions might be semantically equivalent (e.g. \textit{When is easter this year?} and \textit{What day is easter on this year?}) or very close (e.g. \textit{What is the capital of Louisiana?} and \textit{What is the capital of Portugal?}). Similar questions question may be answerable using a different table than the one it is paired with. All questions can be answered via a unique table (i.e. the SQL \verb|JOIN| operator never appears in our dataset). 

A survey led us to distinguish two categories of tables: entity-instance and key-value. An entity instance table contains information about several entities (e.g. Figure 1 is an entity instance table, it contains the presidency information of different presidents in the history of the USA), while a key-value table is related to a single entity (e.g. Figure 2 is a key-value table and it shows some basic information of Donald Trump).

\begin{figure}
  \includegraphics[width=\linewidth]{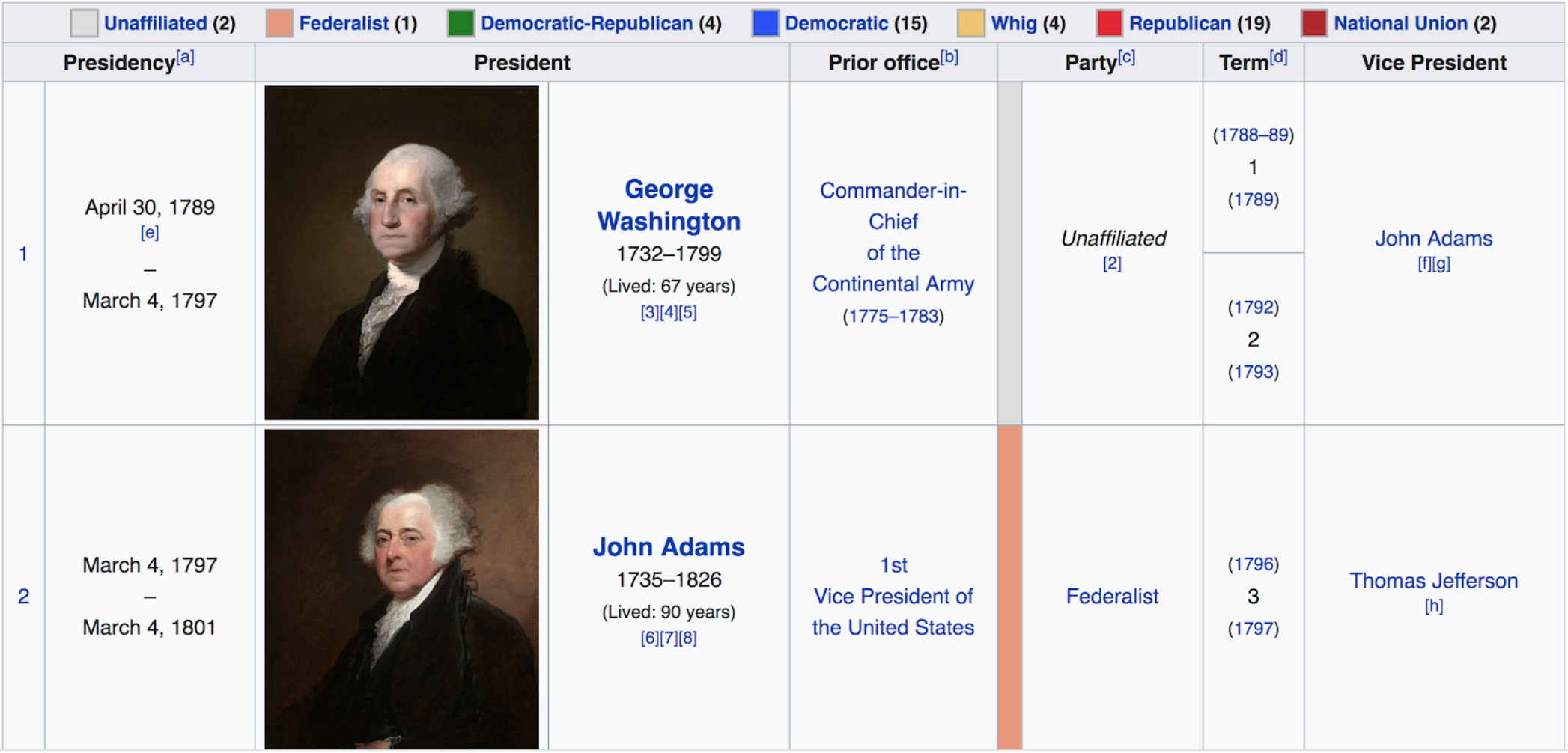}
  \caption{An example of entity-instance table}
  \label{An example of entity-instance table}
\end{figure}
\begin{figure}
  \centering
  \includegraphics[width=0.7\linewidth,height=3.5cm]{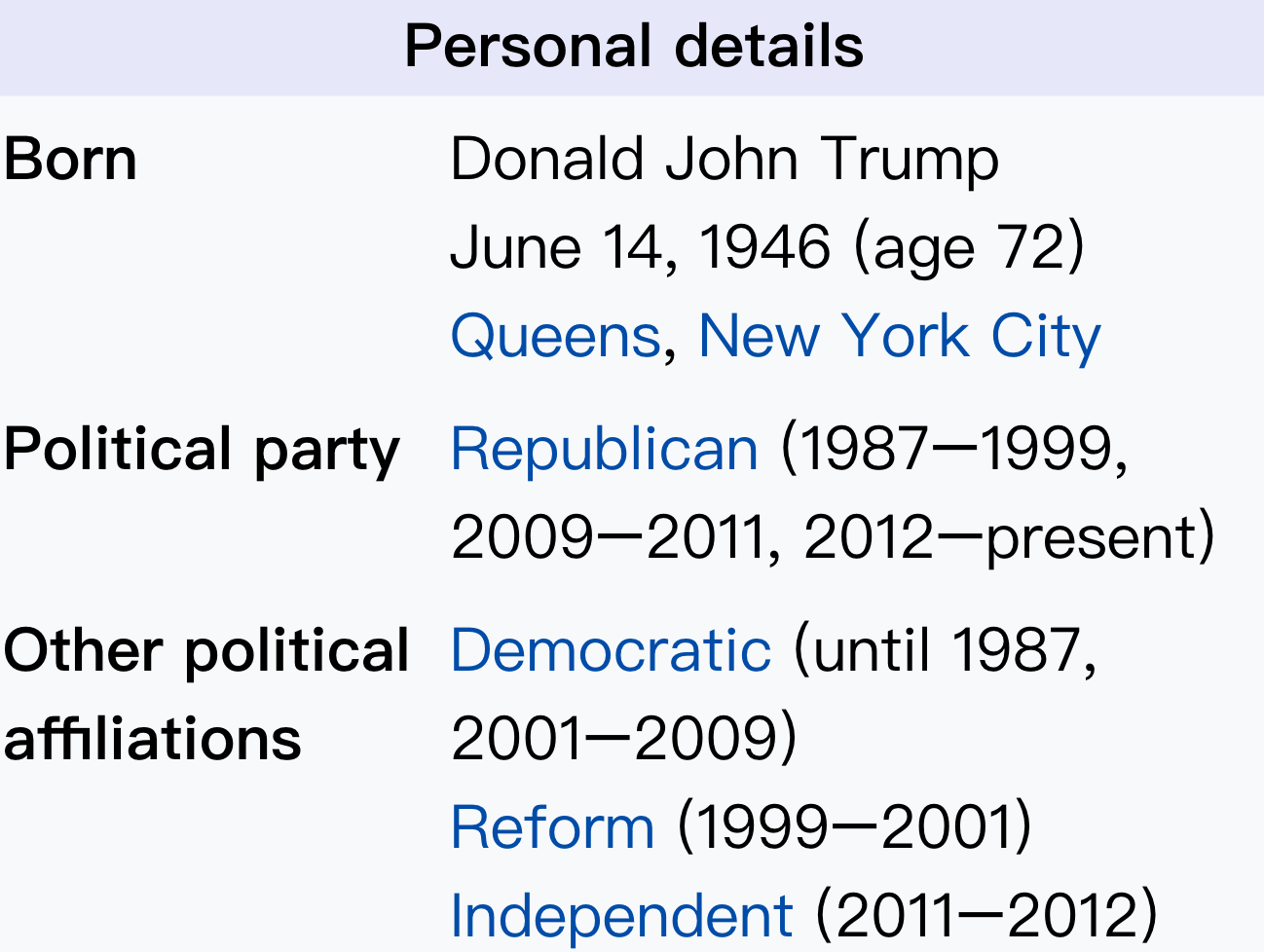}
  \caption{An example of key-value table}
  \label{An example of key-value table}
\end{figure}

\begin{figure*}
    \centering
  \includegraphics[width=0.9\textwidth,height=4cm]{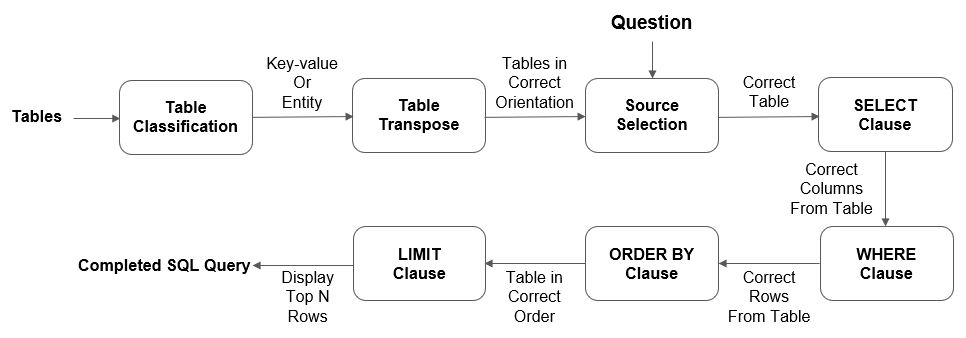}
  \caption{System Design Overview}
  \label{System Design Overview}
\end{figure*}


We removed 64 tables from the full set a wide variety of reasons.

\textbf{Word2Vec limitations}: Occasionally, word embeddings return an incorrect nearest neighbor. For instance, the Query \verb|SELECT| \verb|"President"| \verb|FROM| \verb|"US_Presidents"| \newline \verb|WHERE| \verb|(("President"| $\sim$ \verb|"current president"))|, the nearest neighbor for "president" is "Ulysses S. Grant" instead of "Donald Trump".

\textbf{Ambiguous labelling}: Some tables had answers that had a lack of clarity regarding how the answer could be obtained without recourse to external world knowledge. For the question, \textit {How do I get a refund for Social Security Tax erroneously withheld}, the table had multiple rows on refunds, but not the context to correctly formulate an SQL query that satisfied the labelled answer.

\textbf{Incorrect labelling}: Few of the collected tables did not match with the query, or did not make sense. Occasionally, queries were also formulated incorrectly.

\textbf{Complex and edge cases}: We explain multiple scenarios in detail in Section \ref{sec:Complex cases}, in addition nested queries and those requiring additional operators like AND, OR
did not work well.

\textbf{Badly transposed tables}: Some tables did not allow the answer to be extracted using an SQL query after being transposed (see Section \ref{sec:Design}) due to data loss, non-symmetric tables and data corruption during the operation.


\section{System Design Overview}

Our approach to the question answering problem follows a granular approach. We divide the problem into tasks based on the structure of the SQL query desired. 

\begin{enumerate}
\item {Table Type Recognition}
\item {Table Transpose}
\item {Source Selection}
\item {Complete SELECT Clause}
\item {Complete WHERE Clause}
\item {Complete ORDER BY Clause}
\item {Complete LIMIT Clause}
\end{enumerate}

As shown in Figure \ref{System Design Overview}, we build the SQL query step by step, passing along key information from one step to the next in the pipeline. In the first step, we identify the type of each table, and classify it as either entity-instance or key-value. We then use this information to transpose all the key-value type tables into an equivalent entity-instance type table. For each question, we try to determine which table contains the answer. Once found, we pair the question and table to form the FROM clause for the query. We then extracting which column or columns from the table need to be present in the query, i.e. the SELECT clause of the query. Given the correct columns, we predict the correct the row or rows, i.e. the WHERE clause of the query. This produces a logically correct and complete SQL query for most questions. However, some questions are such that they require to be ordered by a certain column, or require a fixed number of rows as the answer. For example, \textit{Who is the current President of USA?} needs to be ordered by year, and \textit{Who are the top 5 goal scorers in La Liga?} needs a specific number of rows. Hence, the next steps are to determine the columns used for sorting (i.e. the ORDER BY clause of the query) and the number of rows to be displayed (i.e. the LIMIT clause of the query).

After performing all these steps, we obtain all the clauses needed to construct a SQL query which corresponds to the given question and can obtain answers by executing it against the appropriate table. This approach lays bare the reasoning required in each step. However, this approach also propagates error from one step to the next, in an additive fashion.

\section{Experimental Design}
\label{sec:Design}
For each task, we analyzed our dataset, extracted features and built an individual model. We used different metrics to evaluate the performance at every step
\subsection{Machine Learning Models/Algorithms}
\subsubsection{Table Type Recognition}
The type of the table strongly informs the content of the SQL query, so we train different classification models with the following features to classify tables as entity-instance or key-value tables:

\textbf{Number of columns}: In most cases, key-value tables only contain a "key" column and a "value" column. If the number of table columns is equal to 2, then it most likely is a key-value table.  We ignored columns containing only URL in tables for this computation. Columns of this type are a side effect of the process that extracts the table from the web page.

\textbf{"key" or "property" in columns headers}: We observe that most of the key-value tables contain keyword "key" or "property" in the header of "key" column.

\textbf{Normalized variance of content length by number of words}: We first calculate the number of whitespace-separated words within each cell, normalize within the column by the maximum cell length, and determine the variance. For key-value table, the cells within the same column are different attributes of a single entity, leading to a lack of uniform patterns for cell content. Therefore, the cell lengths of the same column tend to be diverse for key-value tables. For entity-instance tables, the content length of each cell in the same column tends to be similar because each column of the entity-instance table corresponds to a specific attribute and it usually has a similar format in the context of the table. If each column in a table has similar length the table is likely to be classified as an entity-instance table.

\textbf{The normalized variance of presence of digits}: Similarly, since each column of entity-instance tables corresponds to a specific attribute, the pattern of the presence of digits would be similar for each cell within the table. As can be seen from Figure \ref{An example of entity-instance table}, the first column is the presidency period, and each cell of that column contains a digit-based description of two years description and two days, so that every cell in that column would contain digits. However, for key-value tables, since cells in the same column correspond to different attributes, the uniform pattern of the presence of digits generally does not hold. If none of the cells or all the cells contain digits for each column, the variance would be zero. The higher the value of this feature is, the more random the presence of digits in a column is. Since the columns of entity-instance tables are more likely to have a uniform pattern of the presence of the digits, a higher variance in the presence of digits is an indicator that the table has a higher probability be a key-value table.

\subsubsection{Tables Transpose}
\begin{table}
\begin{center}
\begin{tabular}{|p{1.6cm}|p{1.6cm}|p{0.8cm}|p{1cm}|p{1.6cm}|}
\hline
\bf spouse & \bf born & \bf height & \bf net worth & \bf education\\
\hline
Melania Trump (m. 2005), Marla Maples (m. 1993-1999), Ivanka Trump (m. 1977-1992) & June 14, 1946 (age 71 years), Jamaica hospital medical center, New York city, NY & 6' 3'' & 3.1 billion USD (2018) & Wharton School of the University of Pennsylvania (1966-1968) ,  more\\
\hline
\end{tabular}
\end{center}
\caption{Key-value table after transposing to entity-instance table}
\label{Key-value table after transposing}

\end{table}
We transpose all key-value tables into entity-instance tables for uniformity. Table \ref{Key-value table after transposing} shows an original key-value table that contains information about Donald Trump that has been transposed into entity-instance. For key-value tables, the goal of the SQL query is to find the target row, while for the transposed entity-instance table, the goal is to find the target column. The ground truth SQL query for the pre-transposed table was "\verb|SELECT| \verb|Value| \verb|FROM| \verb|Donald-Trump| \verb|WHERE| \verb|Key| $\sim$ \verb|birthday|", and the ground truth SQL query for transposed table \ref{Key-value table after transposing} is "\verb|SELECT| \verb|born| \verb|FROM| \verb|Donald-Trump|". All post-transpose algorithms hence only need to consider entity instance tables.

\subsubsection{Source Selection}
To answer a specific question, we first need to correctly identify the source table in the dataset. For question preprocessing, we lowercase all characters and tokenize on whitespace and non-alphanumerics.  We then used NLTK~\cite{bird2009natural} to remove stop words and stem the remaining words. We apply the same method to each tables cell.  We calculate the TF-IDF of word stems in questions and tables. The inverse document frequency (IDF) of each word stem was calculated on all tables, and the term frequency (TF) of each word stem for each question or table was calculated in the scope of that specific question or table.

We explored three different approaches to measure the similarity of the question and the table in the TF-IDF vector space: (1)  Cosine similarity of the two vectors, (2) Dot product of the two vector, and (3) Inverse Euclidean distance of the two vectors (after projected to points in multi-dimensional space). 

\subsubsection{Column Projection in SELECT Clause}
The first choice in composing the SQL query is to determine which columns belong to the SELECT clause. This task can be treated as a binary classification task: given the question and a column in its source table, determine whether this column should be included in the SELECT clause. The input of our classifier are the following features, which are extracted from a pair of question and column to be processed:

\textbf{Number of columns in table}: The number of columns one table contains influences the probability that any column within the table being included in the SELECT clause significantly. If the table only has one column, then the only column must be included in the SELECT clause to construct the SQL query. As the number of columns in the table increases, the probability of each column being included in the SELECT clause decreases.

\textbf{Similarity of column contents and question using word vectors}: We calculate 4 types of proximity: the average proximity of the column contents and the question, the average proximity of the column contents and the question without stop words, the maximum proximity of the column contents and the question, and the maximum proximity of the column contents and the question without stop words. For example, if the question is \textit{Who is the president of the USA?}, the word "president" would be close to "Donald Trump" in word vector space and the column containing "Donald Trump" is exactly the column we want to select.

\textbf{Column Types}: To be able to learn answers for different question categories, we built a model for recognizing the column data type. 
The feature is a 7-dimensional vector with the probability distribution among the 7 column types.

\textbf{Question Types}: We classified questions based on Li et al.'s work\cite{Li:2002:LQC:1072228.1072378} to produce 6 major types : "ABBREVIATION", "ENTITY", "DESCRIPTION", "HUMAN", "LOCATION" and "NUMERIC" and also included yes/no questions as a seventh major type.
In addition, the minor types for "NUMERIC", namely, "date", "count", "period" and "money" were used as the dataset contains many factoid questions. We used the classification API from Madabushi. et.al\cite{Madabushi2016HighAR}. Their API extends Li et al.'s work \cite{Li:2002:LQC:1072228.1072378} and provides usable question type results. 
We augmented these results by heuristic methods and achieve 89.2\% accuracy on the training set and 96.1\% accuracy on the dev set. The mapping for question types and column types are generally intuitive. For example, for questions about the price of some product the column in the SELECT clause should have the "Currency" type. Our final representation has 11 different question types.

\textbf{Semantic similarity of column header and question}: We include a feature measuring the similarity of the column header and the question. We applied same pre-processing technique 
to the column headers. For each word stem in the column header and each word stem in the question, we calculated the character level edit-distance of the two word stems. By surveying our dataset we found that there may be several column headers in the source table that all contain the same question words. For example, a question asks about \textit{"What is NAIRU?"}, and the column headers of the first three columns of the source table are "What is NAIRU? CONCEPTS", "History of NAIRU" and "Different concepts of NAIRU". If we only include the minimum edit distance, then the distance between any of these column headers and the question is equal to zero and the feature is useless. To increase the robustness of our feature, we include both the minimum distance and second lowest distance as features for all word stem pairs.

A concatenation of all features forms a 25-dimensional vector as model input. Our classifier is a multi-layer perceptron (MLP) with 4 hidden layers (ReLu activation, separated by Batch Normalization layers, hidden sizes 32, 16, 8, 2 from bottom to the top). We optimize our model using stochastic gradient descent and cross-entropy loss. Since the number of positive cases is much smaller than the number of negative cases in our train dataset (among the total of 2046 columns, 273 are included in SELECT clause, 1773 are not included in SELECT clause), we duplicated the positive cases 6 times to balance our training data.

\subsubsection{WHERE Clause Projection}
The next component of the SQL query is the WHERE clause, formed by a pair of column and key word. Similar to the task of column projection in SELECT clause, we build a binary classifier which takes the question, one column within the table, and one question word as input. The classifier outputs 1 if and only if the input column and the word both appear in the WHERE clause of the SQL query. We build the following feature vector:

\textbf{Minimum normalized edit-distance between input question word and words within input column}: For the WHERE clause, the input column exploits the input word to constrain the query result to some specific rows within the table. If the input question word appears in the input column, then the probability of this column and question word pair forming a valid WHERE clause increases.

\textbf{Average character-level cell length of input column}: Since it is hard for humans to find key information in extended content, a column with lower average cell length is more likely to appear in the WHERE clause.

\textbf{Number of rows of input column}: If the input column only contains a single row, then we may not need to have a WHERE clause in the SQL query. The classifier in this case would have a higher chance to output 0.

\textbf{Whether the input column has already been included in SELECT clause}: If the input column has already been included in SELECT clause, then it is less likely that it is also used in WHERE clause. We used the ground truth instead of the predicted results for the SELECT clause to build this feature, this decision isolates the prediction errors from the previous model from the current model.

\textbf{Column Types and Question Types}: We used the same column types feature and question types features as we used in SELECT clause task. In SELECT clause task, for one type of question, the probability that one corresponding type of column to be included in SELECT clause is higher than the remaining types. However, for the WHERE clause task, the probability of the column with that specific type appearing in the WHERE clause is relatively low. For example, for "When" type question, it is unlikely that we still use a column with "DateTime" type to filter the query result.

\textbf{POS, NER, and dependency parsing of input word}: Common natural language processing features may give us some clues about the probability of the input word being included in WHERE clause. For parts-of-speech (POS) tagging, some types of words are more likely to be used as keyword in WHERE clause than other types of words. For example, nouns have a higher probability to be selected as the search word in the WHERE clause than adjectives and verbs. Similarly, for named-entity-recogition (NER) tagging, entities such as "PERSON" are more likely to be selected as the search word compared to other entities, like "DATETIME". Moreover, dependency parsing helps us to identify the root word in a sentence, which has a higher chance to be included in the WHERE clause. Therefore, we included POS and NER tagging and dependency parsing of the input word as features in one-hot encodings. There are a total of 12 POS features, 6 NER features and 37 dependency parsing features. 

Using these features, we applied the same MLP model
to the WHERE clause task as used for the SELECt clause task, except that the input size was increased from 25 to 77. We also performed up-sampling to balance our training data. 

\subsubsection{Column Type Recognition}

We pre-defined 7 column types, "DateTime", "Currency", "Percentage", "Numerical", "Boolean", "Text" and "URL". We manually labeled all the columns in the train dataset (total of 2046 columns). Our task is that given the content of a column in the table, determine which type this column belongs to. 

For every column, we extract the following features to cater to specific question types: Proportion of cells that can be directly converted into a valid numerical value; Proportion of cells that contain only digits; Proportion of cells that contain currency symbols, percentage symbols, boolean values (yes, no, true, false), years (1500-2020), month strings and abbreviations, weekday strings, and the `http' string to detect URLs. We build an MLP for column types recognition using 3 hidden layers (ReLu activation, sizes of 32, 32 and 7 from bottom to top), followed by a Softmax layer. The output of the model is a probability distribution among the 7 types given the input column. We assigned the column with the type that has the highest probability. We achieve an accuracy of over 93\% on our manually labeled training data.

\subsection{End-to-end Pipeline}
Given the separate models we have designed for generating the SQL query for a question, we built an end-to-end question answering pipeline using the models we described in the previous section. The input of the pipeline is a question and a collection of tables. The output of the pipeline is a collection of cells which are used for answering the input question extracted from one table within the input collection of tables. The system first retrieves a table using the TF-IDF similarity approach explained above. We then predict the column(s) to be included in the SELECT clause 
as the answer-containing columns. For row selection, we make use of the classifier we designed for WHERE clause.
Since the prediction result of the WHERE clause classifier is whether a pair of column of the table and a key word in the question should be included in the WHERE clause, we need another processing step to map the WHERE clause prediction to rows selection. We designed two different algorithms for this step in our experiment, \textit{Word-Match Based} and \textit{Word2Vec Based}.

\textbf{Word-Match Based}: For each column and key word pair which the classifier predicts as belonging to the WHERE clause, we go over each cell within the column and, if there is a match between the word in the cell and the word in the predicted pair, the row where the cell is located is assigned a score of 1. After all column/word pairs are processed, we select the rows with the highest score. If the classifier predicts that the WHERE clause is empty for the given table, then all rows are selected.

\textbf{Word2Vec Based}: We use pretrained GloVe embeddings~\cite{pennington2014glove} to map the key word and the words in the columns to 300-dimensional vectors. For each cell in the column, we calculate the minimum Euclidean distance between the word vectors in the cell and the vector of the key word. The row where the cell is located is assigned with that minimum distance. After all the pairs are processed, the rows with the global minimum distance are selected. If the WHERE clause is empty, or the key word doesn't exist in the pretrained embeddings, all rows of the table are selected.

\subsubsection{Answer Cell Selection}
After the columns and rows have been selected, their intersecting cells are predicted as the answer. 

\subsection{Evaluation Metrics}
\textbf{Table Type Recognition}: We calculated the accuracy of different machine learning models for table type recognition to evaluate their performance on the classification of the input table as a key-value table or an entity-instance table.

\textbf{Source Selection}: For the source selection task, we use precision at rank k (P@k) as the evaluation metric, in which we calculated the precision of the ground truth source table being selected in the top $N$ most similar tables for a given question, with $N = 1, 3, 5, 10$.

\textbf{SELECT Clause and WHERE Clause}: For the SELECT clause and WHERE clause tasks, since we treat each task as a machine learning classification problem, we use precision, recall, and confusion matrices to measure the performance of our classifiers for train and dev set.

\textbf{End-to-end Pipeline}: For our end-to-end pipeline, we calculated the precision, recall and F1-Score of the selected cells generated by our pipeline with the golden cells for each question.

\section{Experimental Results}

\subsection{Source Selection}
\begin{center}
  \begin{tabular}
  {|p{0.7cm}|p{0.8cm}|p{0.8cm}|p{0.8cm}||p{0.8cm}|p{0.8cm}|p{0.8cm}| }
    \hline
    & \multicolumn{3}{c||}{Train Dataset} & \multicolumn{3}{c|}{Dev Dataset} \\
    \cline{2-7}
    & CosS & DotP & InvEuc & CosS & DotP & InvEuc \\
    \hline
    P@1& 60.2\% & 69.1\% & 70.2\% & 72.3\% & 72.3\% & 72.3\% \\ \hline
    P@3& 78.5\% & 80.1\% & 80.6\% & 85.1\% & 87.2\% & 87.2\% \\ \hline
    P@5& 84.8\% & 84.8\% & 85.9\% & 89.3\% & 91.5\% & 91.5\% \\ \hline
    P@10& 89.0\% & 90.6\% & 91.6\% & 89.3\% & 93.6\% & 93.6\% \\ \hline
  \end{tabular}
\end{center}

We see that Inverse Euclidean distance performs best with an accuracy of 70.2\% on train data and 72.3\% on dev data. After error analysis, we found that some error cases are not strictly wrong due to the fact that multiple sources could be used to answer the same question. Considering these cases, the adjusted P@1 results using Inverse Euclidean distance method are 76.5\% on train and 76.6\% on dev data, respectively.

\subsection{Table Type Recognition}
The table type recognition task is straightforward and we achieve around 98.3\% on train and 100.0\% on dev data across common models (logistic regression, decision trees, KNN).

\subsection{SELECT Clause}



\begin{center}
  \begin{tabular}
  {|p{1.1cm}|p{1.4cm}|p{1.4cm}||p{1.4cm}|p{1.4cm}| }
    \hline
    & \multicolumn{2}{c||}{Train Dataset} & \multicolumn{2}{c|}{Dev Dataset} \\
    \cline{2-5}
    & Predicted:1 & Predicted:0 & Predicted:1 & Predicted:0\\
    \hline
    Actual:1 & \hfil 182 & \hfil 30 & \hfil 38 & \hfil 12  \\ \hline
    Actual:0 & \hfil 209 & \hfil 1001 & \hfil 54 & \hfil 297  \\ \hline
  \end{tabular}
\end{center}

\begin{center}
  \begin{tabular}
  {|p{1.2cm}|p{1.5cm}|p{1.5cm}| }
    \hline
    & \multicolumn{2}{c|}{Test Dataset} \\
    \cline{2-3}
    & Predicted:1 & Predicted:0 \\
    \hline
    Actual:1 & \hfil 28 & \hfil 14\\ \hline
    Actual:0 & \hfil 53 & \hfil 196\\ \hline
  \end{tabular}
\end{center}

\begin{center}
  \begin{tabular}
  {|p{1.2cm}|p{1.4cm}||p{1.4cm}||p{1.4cm}| }
    \hline
    & \multicolumn{1}{c||}{Train Dataset} & \multicolumn{1}{c||}{Dev Dataset} & \multicolumn{1}{c|}{Test Dataset}\\
    \hline
    Accuracy & \hfil 83.2\% & \hfil 83.5\% & \hfil 77.0\%  \\ \hline
    Recall & \hfil 85.8\% & \hfil 76.0\%  & \hfil 66.7\%  \\ \hline
    Precision & \hfil 46.5\% & \hfil 41.3\%  & \hfil 34.6\%  \\ \hline
  \end{tabular}
\end{center}

The tables above present confusion matrices along with accuracy, recall and precision results of predicting the correct columns in the SELECT clause. Our model is able to achieve a good accuracy (83.2\% on training set, 83.5\% on dev set and 77.0\% on test set) and recall (85.8\% on training set, 76.0\% on dev set and 66.7\% on test set) performance, but suffers from low precision. 
Improving precision is a goal for future work. Arguably, for this sub-problem, recall is more important than precision, because we want to make sure we include all the targeted columns in our predicted SQL query so that users will not miss any information in the returned result.

\subsection{WHERE Clause}

\begin{center}
  \begin{tabular}
  {|p{1.1cm}|p{1.4cm}|p{1.4cm}||p{1.4cm}|p{1.4cm}| }
    \hline
    & \multicolumn{2}{c||}{Train Dataset} & \multicolumn{2}{c|}{Dev Dataset} \\
    \cline{2-5}
    & Predicted:1 & Predicted:0 & Predicted:1 & Predicted:0\\
    \hline
    Actual:1 & \hfil 95 & \hfil 0 & \hfil 20 & \hfil 7  \\ \hline
    Actual:0 & \hfil 106 & \hfil 4107 & \hfil 35 & \hfil 1372  \\ \hline
  \end{tabular}
\end{center}

\begin{center}
  \begin{tabular}
  {|p{1.2cm}|p{1.5cm}|p{1.5cm}| }
    \hline
    & \multicolumn{2}{c|}{Test Dataset} \\
    \cline{2-3}
    & Predicted:1 & Predicted:0 \\
    \hline
    Actual:1 & \hfil 16 & \hfil 4\\ \hline
    Actual:0 & \hfil 39 & \hfil 2270\\ \hline
  \end{tabular}
\end{center}

\begin{center}
  \begin{tabular}
  {|p{1.2cm}|p{1.4cm}||p{1.4cm}||p{1.4cm}| }
    \hline
    & \multicolumn{1}{c||}{Train Dataset} & \multicolumn{1}{c||}{Dev Dataset} & \multicolumn{1}{c|}{Test Dataset}\\
    \hline
    Accuracy & \hfil 97.5\% & \hfil 97.1\%  & \hfil 98.2\% \\ \hline
    Recall & \hfil 100.0\% & \hfil 74.1\%  & \hfil 80.0\%  \\ \hline
    Precision & \hfil 47.3\% & \hfil 36.3\%  & \hfil 29.1\%  \\ \hline
  \end{tabular}
\end{center}

From the three tables above, we can see that prediction accuracy for the WHERE clause is very high on both training set (97.5\%) dev set (97.1\%) and test set (98.2\%). However, for recall and precision, the good performance on the train dataset does not generalize to the dev set and test set. This is mainly due to the sparsity of the dataset: not all questions would require a WHERE clause in the converted SQL query. As we can see from the confusion matrix, we only have 120 cases in the train dataset that contain WHERE clause. Hence the model is not able to learn all the general rules to identify the correct "\verb|column| $\sim$ \verb|question_word|" pair in the WHERE clause. 

\subsection{End-to-end Pipeline}
\begin{center}
  \begin{tabular}
  {|p{1.1cm}|p{1.6cm}|p{1.2cm}||p{1.6cm}|p{1.2cm}| }
  \hline
    & \multicolumn{4}{c|}{Table Selection Scope: Golden Table} \\
    \cline{2-5}
    & \multicolumn{2}{c||}{Train Dataset} & \multicolumn{2}{c|}{Dev Dataset} \\
    \cline{2-5}
    & Word Match & Word2Vec & Word Match & Word2Vec\\
    \hline
    Precision & \hfil 37.1\% & \hfil 31.3\% & \hfil 34.7\% & \hfil 25.2\%  \\ \hline
    Recall & \hfil 79.8\% & \hfil 74.1\% & \hfil 72.3\% & \hfil 61.7\%  \\ \hline
    F1 & \hfil 45.4\% & \hfil 38.4\% & \hfil 43.0\% & \hfil 30.8\%  \\ \hline
  \end{tabular}
  \begin{tabular}
  {|p{1.1cm}|p{1.6cm}|p{1.2cm}||p{1.6cm}|p{1.2cm}| }
  \hline
    & \multicolumn{4}{c|}{Table Selection Scope: Individual Set} \\
    \cline{2-5}
    & \multicolumn{2}{c||}{Train Dataset} & \multicolumn{2}{c|}{Dev Dataset} \\
    \cline{2-5}
    & Word Match & Word2Vec & Word Match & Word2Vec\\
    \hline
    Precision & \hfil 26.5\% & \hfil 23.3\% & \hfil 27.1\% & \hfil 19.9\%  \\ \hline
    Recall & \hfil 56.8\% & \hfil 55.6\% & \hfil 51.1\% & \hfil 44.7\%  \\ \hline
    F1 & \hfil 32.6\% & \hfil 29.0\% & \hfil 32.1\% & \hfil 23.3\%  \\ \hline
  \end{tabular}
  \begin{tabular}
  {|p{1.1cm}|p{1.6cm}|p{1.2cm}||p{1.6cm}|p{1.2cm}| }
  \hline
    & \multicolumn{4}{c|}{Table Selection Scope: All Sets} \\
    \cline{2-5}
    & \multicolumn{2}{c||}{Train Dataset} & \multicolumn{2}{c|}{Dev Dataset} \\
    \cline{2-5}
    & Word Match & Word2Vec & Word Match & Word2Vec\\
    \hline
    Precision & \hfil 26.5\% & \hfil 23.4\% & \hfil 21.8\% & \hfil 17.8\%  \\ \hline
    Recall & \hfil 56.3\% & \hfil 55.6\% & \hfil 44.7\% & \hfil 40.4\%  \\ \hline
    F1 & \hfil 32.5\% & \hfil 28.9\% & \hfil 26.5\% & \hfil 21.2\%  \\ \hline
  \end{tabular}
  \begin{tabular}
  {|p{1.5cm}|p{3cm}|p{3cm}|}
  \hline
    & \multicolumn{2}{c|}{Table Selection Scope: Golden Table} \\
    \cline{2-3}
    & \multicolumn{2}{c|}{Test Dataset} \\
    \cline{2-3}
    & \hfil Word Match & \hfil Word2Vec\\
    \hline
    Precision & \hfil 29.5\% & \hfil 21.6\%  \\ \hline
    Recall & \hfil 54.8\% & \hfil 50.0\%  \\ \hline
    F1 & \hfil 32.0\% & \hfil 23.5\%  \\ \hline
  \end{tabular}
  \begin{tabular}
  {|p{1.5cm}|p{3cm}|p{3cm}|}
  \hline
    & \multicolumn{2}{c|}{Table Selection Scope: Individual Set} \\
    \cline{2-3}
    & \multicolumn{2}{c|}{Test Dataset} \\
    \cline{2-3}
    & \hfil Word Match & \hfil Word2Vec\\
    \hline
    Precision & \hfil 24.7\% & \hfil 21.5\%  \\ \hline
    Recall & \hfil 50.0\% & \hfil 45.2\%  \\ \hline
    F1 & \hfil 27.2\% & \hfil 23.3\%  \\ \hline
  \end{tabular}
  \begin{tabular}
  {|p{1.5cm}|p{3cm}|p{3cm}|}
  \hline
    & \multicolumn{2}{c|}{Table Selection Scope: All Sets} \\
    \cline{2-3}
    & \multicolumn{2}{c|}{Test Dataset} \\
    \cline{2-3}
    & \hfil Word Match & \hfil Word2Vec\\
    \hline
    Precision & \hfil 18.4\% & \hfil 15.1\%  \\ \hline
    Recall & \hfil 38.1\% & \hfil 33.3\%  \\ \hline
    F1 & \hfil 19.7\% & \hfil 15.8\%  \\ \hline
  \end{tabular}
 \end{center}
The above table shows the average precision, recall and F1-score of our end-to-end pipeline. We test our pipeline under different experimental environments: (1) `Golden table': we assume the correct table is selected for generating the SQL query; (2) Individual Set: all tables in the training \textit{or} dev \textit{or} test set, respectively, are available for retrieval; (3) All sets: Both training, dev and test set tables are available for retrieval. As expected, with the increase of the table selection scope, the performance of the pipeline decrease. If the incorrect table is selected at the first step, then there is no chance for the pipeline to find the correct cells for the input question. Another thing to note is that the performance of using word match as rows selection method is better than using word2vec as rows selection method. 

\section{Error Analysis}

\subsection{Source Selection}

When performing error analysis for Source Selection, we noticed that some predictions are not strictly incorrect, although the predicted table is different from the actual table as defined in the ground truth. Some pairs of tables contain similar information and could be used to answer the same question. The reasons why we have similar sources in the dataset are due to three main causes: (1) There are some questions that are semantically equivalent, e.g. \textit{When is Easter this year?} and \textit{What day is Easter on this year?}; (2) There are some questions that are in the same type and are very close. e.g. \textit{What is the capital of Louisiana?} and \textit{What is the capital of New Jersey?}; (3) There are some questions that are asking different information about the same thing. e.g. \textit{What time does the Super Bowl start?} and \textit{Where is the Super Bowl being played this year?}. Our experimental results do not account for this redundancy.

\subsection{SELECT Clause}
For a given question, we would like to know whether our model has successfully predicted all the targeted columns in the SELECT clause, and whether our model has only included the targeted columns in the predicted SELECT clause. Among all the 64 cases in the dev data, we found 6 (9.4\%) exact matches between the predicted SELECT clause and the actual SELECT clause. We have another 48 (75.0\%) cases that we include all the required columns but also provide additional columns. We have divided the error cases into six main categories as explained below, all these cases are related to these two types of features to some extent.  
\textbf{Case 1: Question type detected wrongly (7 cases)}: Our question classifier achieves 89.2\% accuracy on the train set and 96.1\% on the dev set. An example misclassification is the question \textit{What is Washington Wizards record?}, which should be identified as "NUMERIC". However, the question classifier detects this question as "ABBREVIATION", therefore, the model would look for "Text" columns instead of "Numeric" columns.

\textbf{Case 2: Column type detected wrongly (3 cases)}: The MLP column type predictor provides a good accuracy performance of around 93\% but, in some cases, due to some special format in the table content, fails to correctly identify the correct column. For example, for the question \textit{How long do cats live?}, the column "Lifespan" is the targeted column which is supposed to be identified as "Numeric" type. However, as this table only has one row, and the content in "Lifespan" column is \textit{"4-5 years (In the wild)"} which has more alphabets than digits, it is identified as a "Text" type.

\textbf{Case 3: Multiple columns in targeted type (24 cases)}: In some cases, it is difficult for the model to distinguish two columns in the same data type and with similar information in the absence of more sophisticated semantic features. For example, for Question \textit{"How many centimeters in an inch?"}, the model selects both "Centimeter" and "Inch" columns in the table. Both these two columns contain numeric data and both column headers appear in the question. Moreover, for some questions even humans may need external information to help us select the correct column. For example, for the question \textit{What is the population of Boston MA?}, the table has two population columns: "2018 Population" and "2016 Population", we would need to know that 2018 is closer to the current year (or make a functionally equivalent temporal assumption).

\textbf{Case 4: Location type detection needed for WHERE-type questions (5 cases)}: One of the limitations in the current column type recognition model is that it is not able to further segregate "Text" type into sub-classes which includes "Location", "Person", "Description", "Instruction" etc. As a result, we are not able to get a high-accuracy performance for WHERE type questions. We intend to remedy this in future work.

\textbf{Case 5: Human Entity type detection needed for WHO-type questions (2 cases)}: Similarly, we would need the ability to identify the "Person" type to correctly answer WHO type questions.

\textbf{Case 6: Binary questions (4 cases)}: The assumption we made regarding the relationship between question types and column types fails for binary questions where the answers are either "yes" or "no". We intend to remedy this in future work.

\subsection{WHERE Clause}
\label{sec:Error WHERE Clause}

In addition to the column-level performance as shown in Section 8.4, we also checked question-level performance on dev dataset: Among the 64 cases in dev dataset, we have 34 (53.1\%) exact matches between the predicted WHERE clause and the actual WHERE clause. We also have another 8 (12.5\%) cases (i.e. Case 1 and Case 2 below) are not considered as wrong because using the predicted WHERE clause also generates the same result as the actual SQL query. After performing error analysis for WHERE clause, we have identified 4 main categories as explained below.

\textbf{Case 1: Single row tables (7 cases)}: For tables that have only a single row, the result would be same whether we include a WHERE clause or not in the SQL query. Most of the single row tables are generated after transposing the key-value tables.

\textbf{Case 2: Slight difference on search keyword (1 case)}:  Sometimes there may be a slight difference between the predicted search keyword in WHERE clause and the actual one. However, our $\sim$ operator
may mitigate this problem. For example, for the question \textit{What is Washington wizards record?}, the WHERE clause in the actual SQL query is \textit{Team $\sim$ Washington wizards}. Our predicted WHERE clause is \textit{Team $\sim$ Washington}. Since in the actual table, the "Team" column only contains "Washington" instead of "Washington wizards", our predicted WHERE clause would return the same result as the one in the actual SQL query.

\textbf{Case 3: External information required (4 cases)}: Some cases would need external information to form the correct SQL query. For example, for the question \textit{Who is the actress that plays Sheldon's mother?} (aside from the implicit focus on the Big Bang Theory TV show) we would need to have the knowledge that Sheldon's mother is Mary Cooper before we could form the correct WHERE clause which is "\verb|Character| \verb|LIKE| \verb|Mary| \verb|Cooper|".

\textbf{Case 4: Incomplete search keyword due to stop-word removal (3 cases)}: A few cases were predicted wrongly due to stop-word removal. For example, for the question \textit{How many feet are in a mile?}, the expected WHERE clause is "\verb|Mile| $\sim$ \verb|a| \verb|mile|", but the predicted WHERE clause is "\verb|Mile| $\sim$ \verb|mile|"  as stop-word "a" is removed.

\subsection{Row Selection: Word Match vs. Embeddings}
In the end-to-end pipeline, we use two methods for answer row selection from the predicted WHERE clause. For word-match based algorithm, errors mainly stem from a failed match between key word in the where clause and the word in the cells. The word in the cells may exist as a synonym of the key word instead of the key word and therefore leads to the failure of the algorithm. 
For word2vec similarity, errors are mainly caused by irregular forms of the key word in the WHERE clause. Through experiment, we found that the key word can be a foreign language word, a number plus a unit, or some other irregular form, all of which do not exist in the pretained word embeddings we use. Overall, this type of semantic mismatch is severe and leads to the word match method performing best.

\section{Discussion}

\textbf{Data Collection}: Table name and column headers were not always available for the tables we collected from the web. In this case, the students collecting the tables were instructed to come up with a table name and/or column headers that was relevant regarding the context (i.e. table, web page). However, the students also knew the question that the table was supposed to answer. The column headers, therefore, could have information ``leaking'' about the answer that has to be returned for a question. When these questions are in the dev set, the gain in accuracy with this leaked information is difficult to gauge. The application scenario assumes that tables are generally extracted independently of the question itself. In future work we plan to tackle this by means of automated column header generators.

\textbf{Table Transpose}: Transposing key-value tables into entity-instance tables provides us with a more homogeneous structure and makes it easier to run Machine Learning models. Most of the transposed key-value tables have only a single row. However, we also have a small number of transposed key-value tables with multiple rows (e.g. comparisons of consumer electronics' specifications with products as columns and features as rows).

\textbf{Source Selection}: Table selection using inverse Euclidean distance achieved 70.2\% P@1 on the train and 72.3\% on the dev dataset. Since we work with similarities based on bag-of-words TF-IDF vector, the model is prone to error from language nuances in the question. Closely related questions where only one word has been changed are often misclassified.

\textbf{SELECT Clause}: Our MLP model for the SELECT clause problem produced a precision of 46.5\% and 41.3\% respectively for train and dev set. The recall of 85.8\% and 76.0\%, respectively, is a good result and provides opportunity for future work in reranking the results using additional features. Errors for the SELECT clause problem arose from multiple sources: (1) misclassification of the question and/or column type, which may be improved by using pre-trained models or clustering; (2) lack of granularity for column type recognition, such as those identifying a location or a human entity; (3) ambiguities that resulted from a table having multiple columns of the same type; (4) yes/no questions which required boolean inference from the predicted cells.

\textbf{WHERE Clause}: For the WHERE clause prediction, precision is most important and our model achieved 47.3\% and 36.3\% for train and dev sets, respectively. Due to the data sparsity issue in our current dataset, it is not able to generalize from training to dev data. Some errors in misclassification of (column, question word) pairs for the WHERE clause problem were on tables that had a single row, which would could be mitigated using a default-row implementation. We use a $\sim$ operator to identify the column that most closely resembles the word in the question we need to consider. Therefore small errors in the selection of the question word will also strictly produce incorrect final answers. We pre-process the question to remove stop words from consideration. However, in some cases, the final result required their use. 

\subsection{Complex and Edge cases}
\label{sec:Complex cases}
There are some complex cases or edge cases need to be further studied after/while we collect more data and refine our model in future work.

\textbf{AND/OR operator}: For some questions, we may need to put AND or OR operators in the converted SQL query. For example, for Question \textit{"What year was Brock Lesnar's last UFC fight?"}, the table contains two (unordered) columns containing contestant names, thus the ground truth SQL query is \verb|SELECT| \verb|"Date"| \verb|FROM| \verb|"Brock Lesnar_Tabe"| 
\verb|WHERE|\newline \verb|(("Fighter_1"| $\sim$ \verb|"Brock Lesnar")| \verb|OR|\newline \verb|("Fighter_2"|  $\sim$ \verb|"Brock Lesnar"))| \verb|AND| \newline\verb|"Event_Name_1"|  $\sim$ \verb|"UFC"| \verb|ORDER BY| \verb|"Date"| \newline
\verb|DESCENDING| \verb|LIMIT 1|

\textbf{Sub query}: For some questions, we would need to use sub-query to generate a complete SQL query to locate the correct answer. For example, for Question \textit{"Who became president after John Kennedy?"}, the ground truth SQL query is \verb|SELECT| \verb|"President"| \verb|FROM|  \verb|"List of Presidents |\newline \verb|of the United States"|
\verb|WHERE| \verb|"Number" >| \newline \verb|(SELECT|
\verb|"Number"| \verb|FROM| \verb|"List of|
\verb|Presidents| \newline \verb|of the United States"|
\verb|WHERE "President"| $\sim$ 
\verb|"John Kennedy")|  \verb|ORDER| \verb|BY| 
\verb|"Number"| \verb|ASCENDING| \verb|LIMIT 1|

\textbf{Aggregate function}: For some questions, we may not be able to find a direct answer in the table. However, we could get the answer using aggregation functions in SQL. For example, for question \textit{"How many science jobs in Rochester NY?"}, the ground truth SQL query, for a table containing science jobs, is \verb|SELECT| \verb|COUNT| \verb|(Title)| \verb|FROM| 
\verb|"Table_1"| \verb|WHERE | \verb|"Location"|
$\sim$ \verb|"Rochester NY"|


\textbf{Questions with answers change over time}:  We also have questions that have ambiguities due to under specification such that the answers change with time. For example, for Question \textit{who won the super bowl?}, the answer would be changed by year. Assuming that the table would be updated so that it contains the most recent record, we have two ways to solve this problem to make sure we always get the most updated answer: (1) We could sort the table by year and get the most recent one. i.e. \verb|ORDER| \verb|BY| \verb|year| \verb|DESCENDING| \verb|LIMIT 1|; (2) We could call an external function which returns the current year from the environment. i.e. \verb|WHERE| \verb|year| \verb|=| \verb|EXTERNAL| \verb|("current| \verb|year")|. The latter approach also applies to questions that require users' location. For example, \textit{"What is the best restaurant near me?"}.

\section{Future Work}
We need to add more questions and tables to the data set. We have observed that in addition to enabling effective use of machine learning techniques, we infer valuable insights from individual tables themselves. Moreover, we also need to have enough training examples to identify the correct features and models to be used for ORDER BY and LIMIT clause problems. A much larger dataset would enable a single large deep neural network that encorporates all the types of reasoning in our current pipeline.

Some question types such as WHERE and WHO have a lower accuracy performance compared to other question types. This poor performance results from low granularity on column type recognition. We need to study how to break down "Text" columns into several different types such as "Location", "Person", "Description", "Instruction", etc.

Although semantic inference problems are not easy to solve, from our error analysis, we know that we are able to further improve the system performance by incorporating more semantic features. This semantic analysis could help us to better answer the binary questions, and also questions like \textit{How many centimeters in an inch?}

An alternative or addition to expanding the data set is using pre-trained models such as those used in \cite{zhong2017seq2sql}.

When we have multiple sources that might answer a question, we should evaluate the top k sources and see if we get the best answer.
We could also explore if a composition of re-ranked results could generate a better answer.

\section{Conclusion}
\label{sec:conclusion}
We proposed to understand the problem of question-answering from structured sources by forming a dataset for modelling and providing
well-defined sub problems. 
In this work, we were able to provide the division for tasks and implement baseline models for each tasks. We also performed error analysis on each task model.

The source selection problems was solved with IR techniques. Both solutions were reasonably accurate and further performance improvements may be gained with fine-tuning.

The SELECT clause problem can be seen as a standard question answering problem that involves mapping question types and column types effectively. We observed limitations with type classification. We also observed problems with disambiguation with multiple matches and semantic inference problems.

The WHERE clause problems proved the hardest as question type knowledge did not have a direct effect on the results. Further, this problem needed both column and the question word to be guessed correctly and therefore the error for the WHERE clause model is higher. 

Our approach by dividing the problem into several well-defined tasks makes it easier to identify the error causes and interpret required areas of focus. After collecting more data and building models for ORDER BY and LIMIT clauses, we would like to complete the framework integration.



\bibliographystyle{abbrv}
\bibliography{sensemaking}


\end{document}